\documentclass[10pt,twocolumn,letterpaper]{article}

\usepackage{wacv}
\usepackage{times}
\usepackage{epsfig}
\usepackage{graphicx}
\usepackage{amsmath}
\usepackage{amssymb}
\usepackage{booktabs}
\usepackage{multirow}
\usepackage{caption}
\usepackage{authblk}
\usepackage[accsupp]{axessibility}

\def\wacvPaperID{273} 

\wacvfinalcopy 

\ifwacvfinal
\def\assignedStartPage{0} 
\fi

\ifwacvfinal
\usepackage[breaklinks=true,bookmarks=false]{hyperref}
\else
\usepackage[pagebackref=true,breaklinks=true,colorlinks,bookmarks=false]{hyperref}
\fi

\ifwacvfinal
\else
\fi

\begin{document}
\title{Dance Style Transfer with Cross-modal Transformer}
\author[*]{Wenjie Yin}
\author[*]{Hang Yin}
\author[$\dag$]{Kim Baraka}
\author[*]{Danica Kragic}
\author[*]{Mårten Björkman}
\affil[*]{KTH Royal Institute of Technology, Stockholm, Sweden}
\affil[$\dag$]{Vrije Universiteit Amsterdam, Amsterdam, Netherlands}
\affil[ ]{{\tt\small yinw@kth.se, hyin@kth.se, k.baraka@vu.nl, dani@kth.se, celle@kth.se}}


\makeatletter
\let\@oldmaketitle\@maketitle
\renewcommand{\@maketitle}{\@oldmaketitle
    \includegraphics[width=\linewidth,height=10\baselineskip]
    {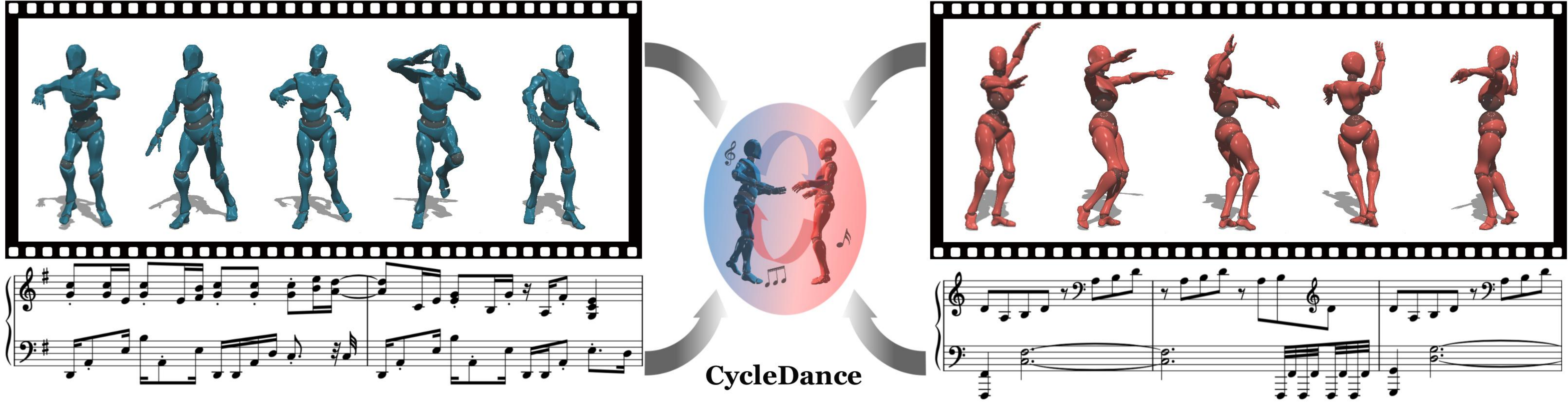}
    \captionof{figure}{Dance style transferred by CycleDance between two dance styles: left) locking dance and right) ballet-jazz dance. The CycleDance framework is trained with unpaired dance motion together with music context.}
    \label{fig:teaser}
    \bigskip}
\makeatother
\maketitle
\thispagestyle{empty}

\begin{abstract}
We present CycleDance, a dance style transfer system to transform an existing motion clip in one dance style to a motion clip in another dance style while attempting to preserve motion context of the dance. 
Our method extends an existing CycleGAN architecture for modeling audio sequences and integrates multimodal transformer encoders to account for music context. We adopt sequence length-based curriculum learning to stabilize training. 
Our approach captures rich and long-term intra-relations between motion frames, which is a common challenge in motion transfer and synthesis work. 
We further introduce new metrics for gauging transfer strength and content preservation in the context of dance movements. We perform an extensive ablation study as well as a human study including 30 participants with 5 or more years of dance experience. 
The results demonstrate that CycleDance generates realistic movements with the target style, significantly outperforming the baseline CycleGAN on naturalness, transfer strength, and content preservation. \footnote{Demo at https://youtu.be/kP4DBp8OUCk.}
\end{abstract}

\section{Introduction}
\thispagestyle{empty}

Style transfer methods facilitate art creation of a target style for media such as images \cite{gatys2016image} and music \cite{brunner2018symbolic}. Similar methods are promising for creators to use an existing dance sequence as a starting point to generate variations across different movement styles. In a video game context, these style variations may -e.g.~be associated to different characters with different attributes or personalities. In a choreographic context, such a tool may lead to hybrid human-artificial creative processes, where style transfers are used to iterate on interesting, unexpected, or complementary variations of an initial choreographic material. 

Existing research on transferring human movement styles largely focuses on simple locomotive or exercise motions~\cite{mason2018few,du2019stylistic,aberman2020unpaired} and domain transfers between adults and children. Technical methods for transferring such sequential data include cycle-consistent adversarial networks (CycleGAN) \cite{zhu2017unpaired} and adaptive instance normalization (AdaIN) \cite{karras2019style}.
However, a research gap remains for applying similar techniques to enable style transfer of more complex movements, such as dance movements. Dance movements usually have no explicit functional purpose and tend to exhibit considerable richness in posture, rhythm and their composition. Generation of dance movements can be particularly challenging since it demands a multi-layer approach that captures motion qualities such as the coordination of joint dynamics and socio-cultural factors associated with the production and perception of the movement. Meanwhile, there exist a variety of such characteristics within different dance styles, originating from different historical backgrounds. Dance styles could be thought more generally of as styles of performing certain dance movements rather than strictly dance genres. This adds another layer of complexity to the generation of high-quality dance movements of a specific target style. 
All these challenges call for computational models that can capture both high-frequency features and long-term dependencies over time, and as such generate realistic dance with aesthetic and coherence.

Moreover, dance is commonly accompanied by music which can provide tremendous clues for understanding and composing movement. Recent works have shown the effectiveness of music-conditioned dance synthesis~\cite{valle2021transflower, chen2021choreomaster}, which can directly generate dance motion given music context. However, it is unclear whether music context will also facilitate style transfer tasks and how such a multi-modal input should be processed in this context. 

In this paper, we propose CycleDance, a multimodal system (see Figure \ref{fig:teaser}) for dance style transfer. CycleDance adopts a generative scheme by extending CycleGAN-VC2 \cite{kaneko2019cyclegan} to work with unpaired data. To tackle the challenges identified above, we exploit a cross-modal transformer architecture \cite{valle2021transflower} that aims to effectively capture relevant features among different modalities so as to enhance style transfer quality. Specifically, we design a two pathways transformer-based architecture to extract temporally aligned motion and music representations in the context of style transfer. We further propose to train CycleDance progressively with a curriculum learning scheme inspired by Fu et al. \cite{fu2021cycletransgan}. This alleviates issues of instability in training large adversarial models and premature convergence that can lead to inferior performance. 
We evaluate our framework on the AIST++ \cite{tsuchida2019aist} dance database, with the analysis focused on transfer between various dance genres.
Two new metrics based on probabilistic divergence and selected key pose frames are proposed to quantitatively assess the quality of dance style transfer. Moreoever, we invite a group of human participants with extensive dancing experience to provide a subjective evaluation and insights from an expert perspective. These evaluations show that CycleDance greatly outperforms a baseline method and its ablative versions.
As an illustration, a video with generated examples can be found at \href{https://youtu.be/kP4DBp8OUCk}{ https://youtu.be/kP4DBp8OUCk}. 


In summary, our contributions are mainly as follows:
\begin{itemize}
\item Our approach is, to the best of our knowledge, the first to combine complex dance motion and music context in the style transfer task, 
unlocking potential applications
in choreography, gaming, and animation, as well as in tool development for artistic and scientific innovations in the field of dance. 
\item We introduce new metrics based on probabilistic divergence and selected key pose frames for gauging transfer strength and content preservation in the context of dance movements. 
\item We provide an extensive user study of the proposed model. The evaluations and insights from a group of experienced dance performers reveal essential aspects of designing such systems.
\end{itemize}

\section{Related Work}
\thispagestyle{empty}

\label{sec:related}
In this section, 
we first provide an overview of prior works on general style transfer in Section \ref{sec:style-transfer} and focus on motion style transfer in Section \ref{sec:motion-style-transfer}. 
As another relevant topic, motion synthesis from multi-modal data will be briefly reviewed in Section \ref{sec:multi-modal}.

\subsection{Style Transfer}
\label{sec:style-transfer}
In recent years, style transfer has achieved impressive progress in computer vision, speech processing, music processing, natural language processing, motion animation, etc. 
In computer vision, the pioneering work of Gatys et al. \cite{gatys2016image} introduces the concept of style transfer and uses convolutional neural networks (CNNs) to merge the style and content between arbitrary images. 
Huang et al. \cite{huang2017arbitrary} propose an adaptive instance normalization (AdaIN) layer to directly apply arbitrary target styles to an image.  
Zhu et al. \cite{zhu2017unpaired} propose CycleGAN, using a pair of generators and discriminators to translate image style. 
The general idea of CycleGAN has been further developed and improved in StarGAN \cite{choi2018stargan}, with domain labels as additional input, so that image styles can be transformed to multiple corresponding domains, such as facial appearances and expressions.

In research on voice conversion (VC), Kaneko and Kameoka \cite{kaneko2017parallel} develop CycleGAN-VC based on CycleGAN, but with gated CNNs and an identity-mapping loss. 
This is further improved by CycleGAN-VC2 \cite{kaneko2019cyclegan} which adopts two-step adversarial losses, a 2-1-2D convolution structure, and PatchGAN. 
Fu et al.~\cite{fu2021cycletransgan} further incorporate transformers and curriculum learning in voice conversion. 
Research has also been conducted to transfer symbolic music styles, with examples such as 
Groove2Groove \cite{cifka2020groove2groove}, which employs an encoder-decoder structure and parallel data, 
and \cite{brunner2018symbolic} for MIDI music with a CycleGAN-based approach. 

For style transfer in natural language processing (NLP), Mueller et al.~\cite{mueller2017sequence} propose recurrent variational auto-encoders (VAE) to revise text sequences. 
Fu et al.~\cite{fu2018style} construct a multi-decoder and a style-embedding model to learn independent content and style representations with adversarial networks. 
Dai et al.~\cite{dai2019style} propose a Style Transformer network with a special training scheme, which employs an attention mechanism and makes no assumption about the latent representation. 
%

Our work focuses on transferring motion data, in particular, dance movements. We adopt CycleGAN-VC2, previously used for voice conversion, as the basis of our framework and augment training with an extra music modality.

\subsection{Motion Style Transfer}
\label{sec:motion-style-transfer}

Early works on motion style transfer rely on hand-crafted features \cite{amaya1996emotion, unuma1995fourier, witkin1995motion, aristidou2017emotion, hsu2005style}, while
most modern studies advocate learning by extracting features from data \cite{holden2016deep, du2019stylistic, holden2017fast, smith2019efficient, li2021ai, mason2022real, park2021diverse, dong2020adult2child}. Typical models in use include convolutional auto-encoders  \cite{holden2016deep},  CycleGAN \cite{dong2020adult2child}, temporal invariant AdaIN layers \cite{aberman2020unpaired}, autoregressive flows \cite{wen2021autoregressive} and spatial-temporal graph neural networks \cite{park2021diverse}. Some research also concerns efficient generation for real-time style transfer \cite{xia2015realtime, smith2019efficient, mason2022real}.
All these works target relatively simple human movements, such as locomotion and exercise, for which the variation in style is often limited e.g. transferring between adult and children locomotion~\cite{dong2020adult2child}.

Our work handles transfer of dance movements that exhibit substantial richness in terms of postures, rhythms, transitions and artistic styles and as such may be of greater empirical value for e.g. video game or film industries. To handle such complexities, our method significantly differs from the reviewed work,
with transformer and curriculum learning leveraged on top of CycleGAN-VC2 for more effective training on more complex movement data.

\subsection{Music-conditioned Motion Synthesis}
\label{sec:multi-modal}

A plethora of research works have focused on human motion synthesis \cite{butepage2017deep, yan2019convolutional, habibie2017recurrent, yin2021graph, li2017auto}. 
Since dance is often combined with music, cross-modal motion generation, an emerging research topic that explores the association between different modalities, is often explored for better understanding of human motion and music-conditioned motion synthesis.
Most early works focus on statistical models \cite{shiratori2006dancing, fan2011example, lee2013music} and typically generate motions by selection. To be specific, this means synthesizing motion by selecting the motion segments in a database whose features (such as rhythm, structure, and intensity) match each music segment.  
With the development of deep learning, learning-based methods have also been explored. 

For example, in ChoreoMaster \cite{chen2021choreomaster}, an embedding module is designed to capture music-dance connections. 
Sun et al.~\cite{sun2020deepdance} propose DeepDance, a cross-modal association system, which correlates dance motion with music, and 
Lee et al.~\cite{zhuang2020music2dance} a decomposition-to-composition framework that leverages MM-GAN for music based dance unit organization. 
In DanceNet \cite{zhuang2020music2dance}, a musical context aware encoder fuses music and motion features, while
in DanceFormer \cite{li2021dancenet3d} kinematics enhanced transformer guided networks are used for motion curve regression. 
More recently, cross-modal transformers have been successfully applied to model the distribution between music and motion \cite{valle2021transflower}. 

Music-conditioned dance synthesis aims to generate dance motion sequences associated to a given music context. Our work explores the dance style transfer task, focusing on manipulating the style of the existing dance movements while keeping the contextual information. The music modality is not mandatory for our style transfer model to work, but can be incorporated to benefit the generation quality when the data is available.

\section{Methodology}
\thispagestyle{empty}

\label{sec:method}

This section formulates our target problem and establishes notations used throughout the paper. Preliminaries about CycleGAN and CycleGAN-VC2 are also given for self-containment. On the basis of these, we present the contributed technical framework CycleDance.

\begin{figure*}
  \includegraphics[width=\textwidth]{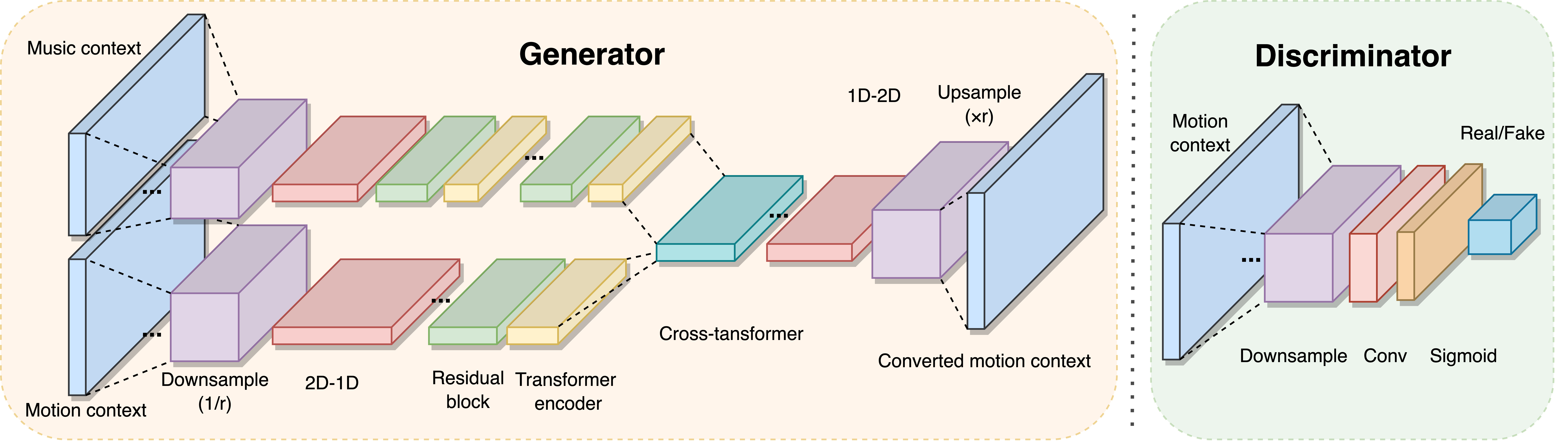}
  \caption{The CycleDance architecture. In the generator, there is a motion pathway and a music pathway. Each pathway starts with downsampling blocks, followed by a 2D-1D block. The motion, music and cross-modal transformer blocks are standard full-attention transformer encoders. Then the fused path is followed by a 1D-2D block and upsampling blocks. In the discriminator, like in Kaneko et al. \cite{kaneko2019cyclegan}, convolution at used in the last layer. }
  \label{fig:model}
\end{figure*}

\subsection{Problem Formulation}

\begin{figure}[tp!]
  \centering
  \includegraphics[width=\linewidth]{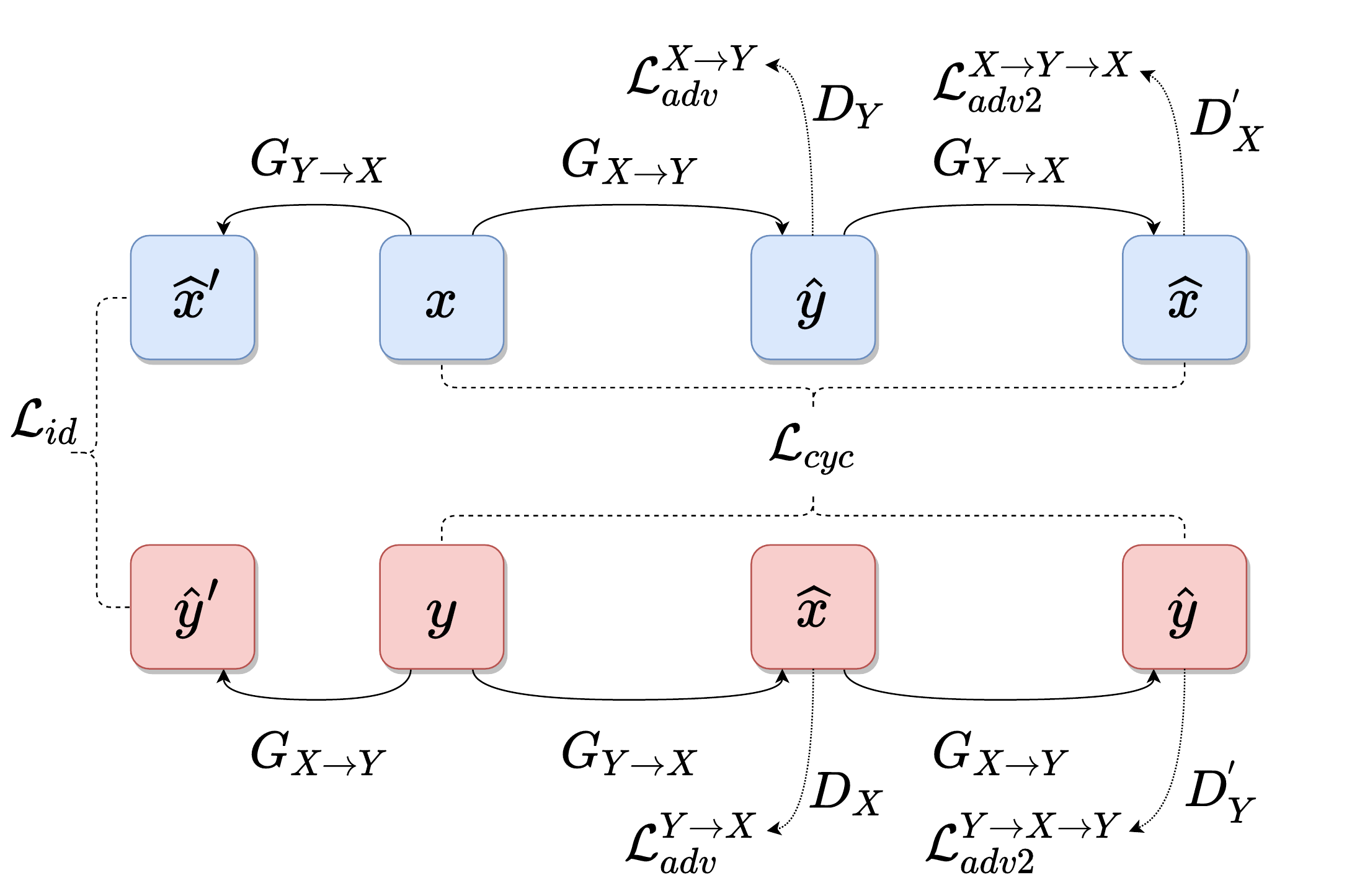}
  \caption{The two step adversarial generative training strategy. The full objective includes four types of losses; adversarial loss $\mathcal{L}_{adv}$, cycle-consistency loss $\mathcal{L}_{cyc}$, identity-mapping loss $\mathcal{L}_{id}$, and second adversarial loss $\mathcal{L}_{adv2}$. See Section \ref{sec:strategy} for the definition of notations. } \label{fig:strategy}
\end{figure}

Our goal is to learn mapping functions between two domains $X$ and $Y$ without relying on paired data between these domains. In our scenario, 
we transfer dance between two style domains  $X$ and $Y$ given dance sample $x\sim P_X$ and $y\sim P_Y$. The dance samples may be paired with music $m_x\in M_x$ and $m_y\in M_y$ with associated styles, although the music modality is only optional in the transfer task.
%

\subsection{Adversarial Training Loss and Strategy}
\label{sec:strategy}
We address the formulated problem with a CycleGAN-like architecture \cite{zhu2017unpaired}, 
as illustrated in Figure \ref{fig:strategy}. The architecture includes  two discriminators $D_X$ and $D_Y$ which are used to distinguish the real and generated data, as well as two mappings $G_{X\rightarrow Y}$ and $G_{Y\rightarrow X}$ for generating patterns of the target style. The mappings are also cycled such that the generated patterns can be converted back to the original domains.
To this end, we follow CycleGAN-VC2 \cite{kaneko2019cyclegan} and incorporate four types of losses, also see Figure \ref{fig:strategy}.

\textbf{Adversarial loss} $\mathcal{L}_{adv}^{X\rightarrow Y}$: this loss measures the discrepancy between the transferred data $G_{X\rightarrow Y}(x, m_x)$ and the target $y$, with the discriminator $D_Y$ attempts to distinguish the transferred data from real data:
\begin{equation}
\label{eq:adv1}
\begin{aligned}
\mathcal{L}_{adv}^{X\rightarrow Y}& = \mathbb{E}_{y\sim P_Y}[\log D_Y(y)]\\
& + \mathbb{E}_{x\sim P_X}[\log(1- D_Y(G_{X\rightarrow Y}(x, m_x)))].
\end{aligned}
\end{equation}
%
Correspondingly, the adversarial loss $\mathcal{L}_{adv}^{Y\rightarrow X}$ can be defined for $G_{Y\rightarrow X}$ and discriminator $D_X$. 
%

\textbf{Cycle-consistency loss} $\mathcal{L}_{cyc}$: this accounts for the loss of contextual information by recovering the original $x$ and $y$ from generated patterns through $G_{X\rightarrow Y}(x, m_x)$ and $G_{Y\rightarrow X}(y, m_y)$: 
%
\begin{equation}
\label{eq:cyc}
\begin{aligned}
\mathcal{L}_{cyc}& = \mathbb{E}_{x\sim P_X}[\left \| G_{Y\rightarrow X}(G_{X\rightarrow Y}(x, m_x))-x \right \|_1]\\
& + \mathbb{E}_{y\sim P_Y}[\left \| G_{X\rightarrow Y}(G_{Y\rightarrow X}(y, m_y))-y \right \|_1].
\end{aligned}
\end{equation} 
%
%

\textbf{Identity-mapping loss} $\mathcal{L}_{id}$: this further encourages input preservation by enforcing an identity transformation when $G_{X\rightarrow Y}$ and $G_{Y\rightarrow X}$ are applied to the other domain:
%
\begin{equation}
\label{eq:id}
\begin{aligned}
\mathcal{L}_{id}& = \mathbb{E}_{x\sim P_X}[\left \| G_{Y\rightarrow X}(x, m_x)-x \right \|_1]\\
& + \mathbb{E}_{y\sim P_Y}[\left \| G_{X\rightarrow Y}(y, m_y)-y \right \|_1]
\end{aligned}
\end{equation}
%

\textbf{Two-step adversarial loss} $\mathcal{L}_{adv2}$: this is a second adversarial loss to alleviate the over-smoothing reconstruction statistics in the cycle-consistency loss~\cite{kaneko2019cyclegan}. Note this introduces
an additional discriminator $D_{X}^{'}$ and $\mathcal{L}_{adv2}^{Y\rightarrow X\rightarrow Y}$ can be defined in a similar way:
%
\begin{equation}
\label{eq:adv2}
\begin{aligned}
& \mathcal{L}_{adv2}^{X\rightarrow Y\rightarrow X} = \mathbb{E}_{x\sim P_X}[\log D_{X}^{'}(x)]\\
& + \mathbb{E}_{x\sim P_X}[\log(1- D_{X}^{'}(G_{Y\rightarrow X}(G_{X\rightarrow Y}(x, m_x))))]
\end{aligned}
\end{equation}
%

The overall objective is finally written as a weighted sum of the above terms
\begin{equation}
\label{eq:full}
\begin{aligned}
\mathcal{L}_{full}& = \mathcal{L}_{adv}^{X\rightarrow Y} + \mathcal{L}_{adv}^{Y\rightarrow X} + \lambda_{cyc}\mathcal{L}_{cyc} + \lambda_{id}\mathcal{L}_{id} \\
&+ \mathcal{L}_{adv2}^{X\rightarrow Y \rightarrow X} + \mathcal{L}_{adv2}^{Y\rightarrow X \rightarrow Y},
\end{aligned}
\end{equation}
where $\lambda_{cyc}$ and $\lambda_{id}$ trade off the consistency and identity loss terms.

Besides, we adopt a curriculum learning algorithm as the training scheme. The intuition is that the training can be more effective by starting with simpler data and progressively handle more complex data.
Such a strategy has been applied to various applications and scenarios, showing an ability to improve the convergence rate and generalization capacity, and providing better training stability \cite{wang2021survey}. 
We adopt a length-based curriculum learning strategy by training data truncation, similar to \cite{fu2021cycletransgan}. The length of input sequences is increased gradually to allow the model to learn from short samples to long samples.

\subsection{Network Architecture}
\thispagestyle{empty}

Our CycleDance framework adopts CycleGAN-VC2 as the backbone  and extends it with a cross-modal transformer, as depicted in Figure \ref{fig:model}.
The cross-modal transformer concatenates two pathways of motion and music encodings, both of which are obtained through a sequence of layers including 2D convolution (purple blocks in Figure \ref{fig:model}), 2D-1D reshaping (red), residual convolution (green) and modality-specific transformers (yellow). The 2D CNN layers are used to perform downsampling while preserve the original sequential structure. The downsampled features are reshaped and pass through the residual blocks of 1D CNNs. The reshaped 1D sequences are processed by transformers which adopt a position embedding and output encodings capturing temporal relations among timesteps. In these blocks, we adopt gated linear units (GLUs) \cite{dauphin2017language} as a tunable activation function to learn a sequential and hierarchical structure. 
%
%

%

For the discriminator, CycleDance also first downsamples motion data with a 2D CNN. 
We only use convolution at the last layer to alleviate training instability, as is suggested in \cite{kaneko2019cyclegan}. The output layer uses sigmoid activation to predict whether the motion clip is real or generated. 



\section{Experiments and Evaluations}
\label{sec:exp}
\thispagestyle{empty}

In this section, we first describe the used dataset (Section \ref{sec:data}), how they are processed and the concrete experimental setup. 
We then detail our assessments, including both objective (Section \ref{sec:objeva}) and subjective (Section \ref{sec:subeva}) evaluations,  and report their results on benchmarking different dance style transfer methods and ablations.

\subsection{Dataset}
\label{sec:data}
We generate 3D dance motion samples with paired music from an existing database called AIST++ Dance Database \cite{li2021ai}. 
AIST++ reconstructs 3D motions from multi-view videos in the AIST Dance Database in terms of SMPL parameters \cite{tsuchida2019aist}. 
%
%
To obtain motion features, we downsample all the motion data to 30 frames per second (fps) and retarget the motion to a skeleton with 21 body joints using Autodesk MotionBuilder. 
Similarly to \cite{valle2021transflower}, we use exponential map parametrization of the 3D rotation to represent all the joints (non-root). The root joint (hip) has four extra features representing the vertical root position, the ground-projected position changes, and the 2D facing angle changes. In total, each motion frame of dance sequences is represented by a 63-dim vector.
The music features are extracted with the Librosa toolbox in a similar way to \cite{tsuchida2019aist}. We combine 20-dim MFCC, 12-dim chroma, 1-dim one-hot peaks, and 1-dim one-hot beats, resulting in a total 35-dim audio feature. 

The selected dance styles are ballet-jazz, locking, waacking, hip-hop, pop and house dance. As the example of data statistics, the ballet-jazz set and the locking dance set both contain 141 motion sequences with 6 songs, lasting 1910.8 seconds and 1898.5 seconds respectively. 

\subsection{Baseline Models and Ablations}

We implement the proposed CycleDance model and the CycleGAN-VC2 baseline  with PyTorch and train both models on the preprocessed dataset. 
In order to assess the significance of contributed design choices, such as the cross-modal transformer and curriculum learning strategy, three alternative architectures are also implemented for an ablation study. 
%
%
The first ablated configuration is CycleTransGAN, which removes the music pathway and the cross-modal transformer and disables the curriculum learning strategy. We expect to use this comparison to highlight the utility of the introduced transformer architecture. 
The second ablation, CycleTransGAN+CL, applies curriculum learning to CycleTransGAN. We aim to assess the performance gains by meticulously modulating the complexities of samples that the model is exposed to during training.   
The final ablation, CycleCrossTransGAN, also uses cross-modal transformers for motion and music information as the encoder. Curriculum learning is, however, not adopted in this configuration. We aim to see the impact of having cross-modal transformers by analyzing the differences between CycleTransGAN and CycleCrossTransGAN. 

\subsection{Objective Evaluation}
\label{sec:objeva}

\begin{table*}[]
\caption{\textbf{Quantitative objective evaluation: }Motion Fréchet distance (MFD) and pose Fréchet distance (PFD) for the five competing models, includes the baseline model, our proposed CycleDance, as well as the ablations. BJ2LC denotes transferring from ballet-jazz to locking dance. Correspondingly, LC2BJ denotes transferring from locking dance to ballet-jazz. Similarly, WK, HP, PO, HO denotes waacking, hip-hop, pop, and house dance. 
}
\resizebox{\linewidth}{!}{
\begin{tabular}{ccccccc|cccccc}
\toprule
\multirow{2}{*}{Method} & \multicolumn{6}{c|}{MFD}                                                                                  & \multicolumn{6}{c}{PFD}                                                                                   \\
                        & BJ2LC           & LC2BJ           & WK2HP           & HP2WK           & PO2HO           & HO2PO           & BJ2LC           & LC2BJ           & WK2HP           & HP2WK           & PO2HO           & HO2PO           \\ \midrule
CycleGAN-VC2            & 9.9430          & 3.4063          & 1.4354          & 1.2645          & 2.2841          & 1.9515          & 0.4897          & 0.3499          & 0.4847          & 0.3313          & 0.5212          & 0.5625          \\
CycleTransGAN           & 3.5643          & 0.7886          & 1.0564          & 0.9464          & 1.5515          & 1.5354          & 0.4749          & 0.2501          & 0.4754          & 0.2834          & 0.4048          & 0.5215          \\
CycleTransGAN+CL        & 2.9188          & 0.5848          & 1.0847          & 0.9847          & 1.4852          & 1.5521          & 0.4897          & 0.2543          & 0.4644          & 0.2882          & 0.4125          & 0.4185          \\
CycleCrossTransGAN      & 2.7446          & 0.5819          & 0.9872          & 1.0782          & 1.4254          & 1.5251          & 0.4419          & 0.2244          & 0.4490          & 0.2880          & 0.3841          & 0.4126          \\
\textbf{CycleDance}     & {2.6109} & {0.5755} & {0.8752} & {0.9501} & {1.3452} & {1.4855} & {0.4216} & {0.2230} & {0.4485} & {0.2960} & 0.3954 & 0.3827 \\ \bottomrule
\end{tabular}
}

\label{tab:obj}
\end{table*}

The main task of all these models is to transfer dance style from a source to a target dance style. To allow for thorough quality assessment of complex motion patterns, common in dance, we perform evaluations from both objective and subjective perspectives.
In the objective evaluation, we use 17 dance sequences per style. We transferred the style for each ablated model and evaluated two metrics that capture how well the style is transferred and how well the content is preserved. To this end, we design metrics based on the Fréchet distance, similar to \cite{valle2021transflower}. 

\textbf{Transfer strength}. The most important aspect of style transfer is transfer strength, which measures the degree of conversion from the source style to the target style. To assess the transfer strength, for one dance style, we use the Fréchet distance between the true dance motion and the generated dance motion. Specifically, we use two consecutive raw poses $(x_{i-1}, x_{i})$ to convert the representations of both true and generated motions to joint velocity $v_i$, without normalization. Similarly, we use three consecutive poses $(x_{i-1}, x_{i}, x_{i+1})$ to calculate the joint acceleration $a_i$. We call this measure the motion Fréchet distance (MFD) and use it to measure how close the generated motion is to the true motion of a target style. 

\textbf{Content preservation}. Another indispensable evaluation metric for style transfer is content preservation. For this dimension, for one dance movements, we use the Fréchet distance between distributions of key poses $x_k$. Keyframes containing such poses are extracted by detecting local maxima in joint acceleration. To make frames comparable, skeleton poses in keyframes are normalized to a hip-centric origin. We call this measure the pose Fréchet distance (PFD) and evaluate to what extent these salient poses are kept after the transfer.


Table \ref{tab:obj} summarises the quantitative results of the proposed model and ablations. 
We transfer the dance style between three pairs of dance genres in both directions, including 'ballet-jazz to locking dance' (BJ2LC) and 'locking dance to ballet-jazz' (LC2BJ), 'waacking to hip-hop dance' (WK2HP) and 'hip-hop to waacking dance' (HP2WK), as well as 'pop to house dance' (PO2HO) and 'house to pop dance' (HO2PO).

\begin{figure}[ht]
  \includegraphics[width=\linewidth]{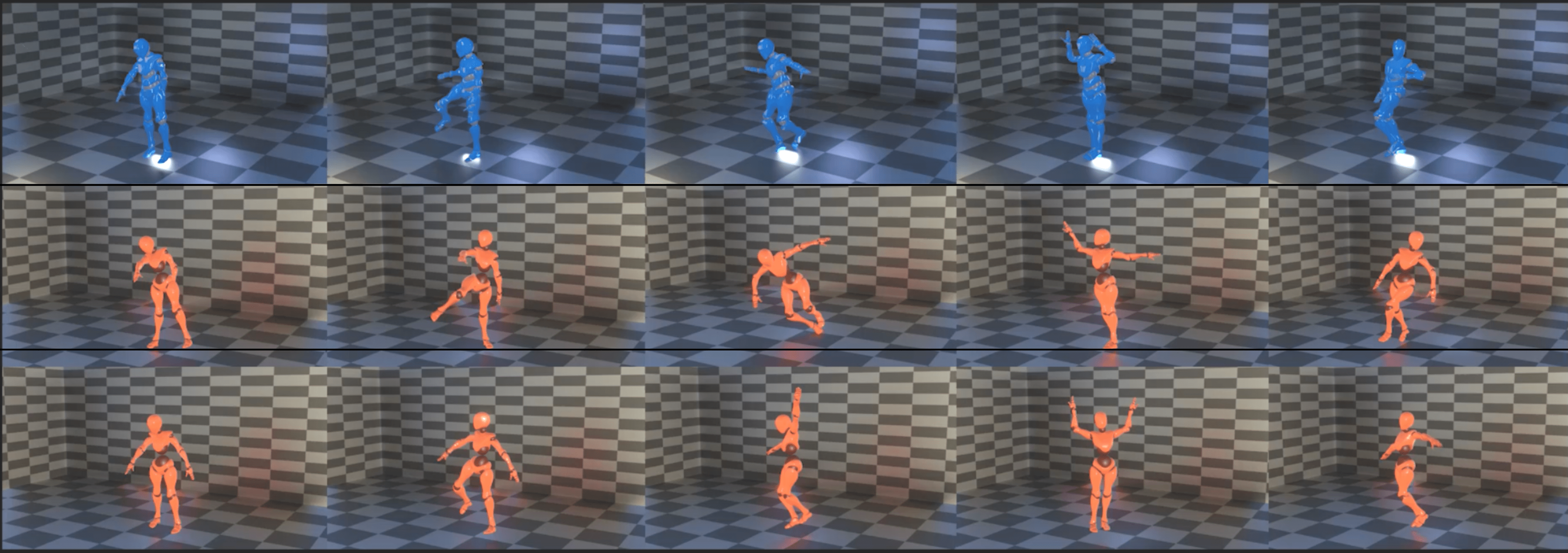}
  \caption{Example locking dance sequences (top, blue y-bot) transferred to ballet-jazz dance by CycleGAN-VC2 (mid, red x-bot) and CycleDance (bottom, red x-bot). }
  \label{fig:obj2}
\end{figure}

\begin{figure}[ht]
  \includegraphics[width=\linewidth]{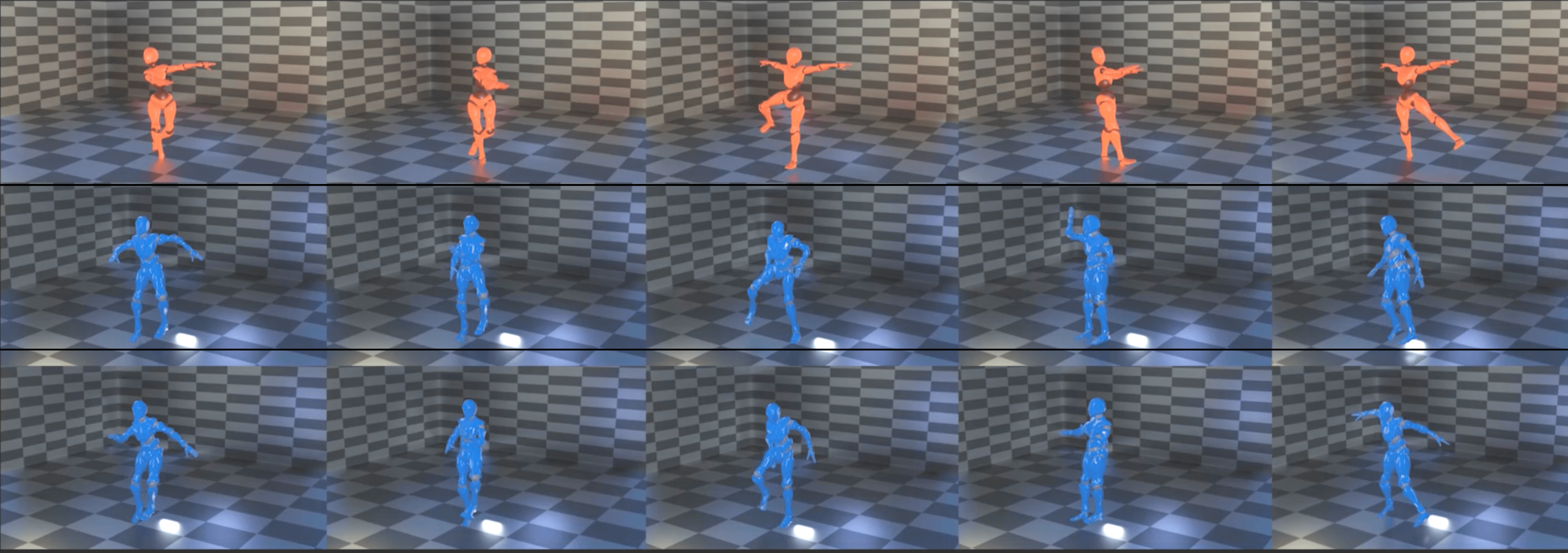}
  \caption{Example ballet-jazz dance sequences (top, red x-bot) transferred to locking dance by CycleGAN-VC2 (mid, blue y-bot) and CycleDance (bottom, blue y-bot). }
  \label{fig:obj3}
\end{figure}

We observe that the baseline model CycleGAN-VC2 struggles in this style transfer task, evident from the much higher MFD for the baseline model than for all other ablation methods. The complete framework, CycleDance, achieves the best performance on both metrics and almost all transfer pairs. This highlights the necessity of all introduced techniques in this task. 
%

An example of synthesized motion clip is presented in Figure \ref{fig:obj2}, which illustrates dance style transfer from locking to ballet-jazz dance. The top keyframe sequence shows the original locking dance. The sequence in the middle is generated by CycleGAN-VC2 and the bottom one by the proposed CycleDance. 
Another example in Figure \ref{fig:obj3} shows dance style transfer from ballet-jazz to locking dance. 
By comparing the poses of each column, it can be observed that the extracted key gestures are representative to the pose sequences.  CycleDance has a higher similarity to the source gestures and can thus preserve more content while having better alignment to the target dance style. 

In addition, through the ablation study, we observe that CycleTransGAN (CycleGAN-VC2 and transformer combined) achieves lower MFD, which can be seen as, with the help of the transformer, the model benefits from capturing richer intra-relations among frames. 
By comparing CycleTransGAN and CycleCrosTransGAN, both MFD and PFD are improved. We take this as evidence that the music information facilitates accurate generation of the target style and that this context information is successfully encoded by the cross-modal transformer. 
The comparison between of CycleTransGAN and CycleTransGAN+CL reveals that curriculum learning greatly improves transfer strength, showing the effectiveness of gradually increasing the difficulty by training with longer clips. 
%

\subsection{Subjective Evaluation}
\label{sec:subeva}
\thispagestyle{empty}

In addition to the objective evaluation, we conducted a user study to evaluate our model and the baseline by scoring three aspects: motion naturalness, transfer strength, and content preservation. We also ask some open-ended questions to gather opinions that may not be covered by the above aspects,
to provide suggestions for future work. 

Our analysis mainly focuses on ballet-jazz and locking dance, since the characterics of these are well understood by dance professionals. 
The user study was performed through an online survey covering transfer tasks for both 'locking dance to ballet-jazz' and 'ballet-jazz to locking dance'. We used Blender to render 8-second video clips with an x-bot character (for ballet-jazz) and a y-bot character (for locking dance) for each source and target dance sequence. 
The participants could play video clips and get acquainted with the animated dance in an introduction phase. In the actual survey, the participants were asked to watch a source dance video clip and a corresponding generated target dance clip. The target dance video clip was generated either from CycleDance or from the baseline, and the order of target dance clips was randomly selected and balanced to relieve potential order effects.  The participants could repeatedly play the clips before answering three questions:
%
\begin{itemize}
\item \textbf{Motion naturalness}: 
\textit{ To what extent do you agree with the following statement? --- The generated motion clip looks natural after style transfer.} (Likert item ranging from 1 (strongly disagree) to 5 (strongly agree)). 

\item \textbf{Transfer strength}: 
\textit{ To what extent do you agree with the following statement? --- The generated motion clip looks like the target dance style. } (Likert item ranging from 1 (strongly disagree) to 5 (strongly agree)).

\item \textbf{Content preservation}:  
\textit{Which feature(s) do you think is (are) the most preserved between the original and the result video? --- Orientation through space; --- Shapes of the limbs; --- Shape of the body trunk; --- Rhythmic patterns --- Other: \_\_\_}. (One or more of these four aspects could be selected). This list was based on the most salient features that dance analysts look at when analyzing expressive movement~\cite{newlove2004laban}.
\end{itemize}

\thispagestyle{empty}

In the study, 30 participants with at least 5 (cumulative) years of dance experience (including training, performing, choreographing, or teaching) were recruited. 
Participants were between 20 and 41 years of age (median 30), 37.9\% male, 58.6\% female, and 3.4\% others.
According to the demographic questions, the participants' familiarity with the ballet-jazz dance and locking dance were M=3.93 (SD=1.05) and M=3.03 (SD=1.18) respectively, where 1 meant not at all and 5 meant very familiar. 
Since the generated motions were shown using virtual characters, we also counted the frequency at which participants played video games, which were 34.5\% weekly, 13.8\% monthly, 13.8\% yearly, and 37.9\% rarely.

\begin{figure}[tp!]
  \centering
  \includegraphics[width=\linewidth]{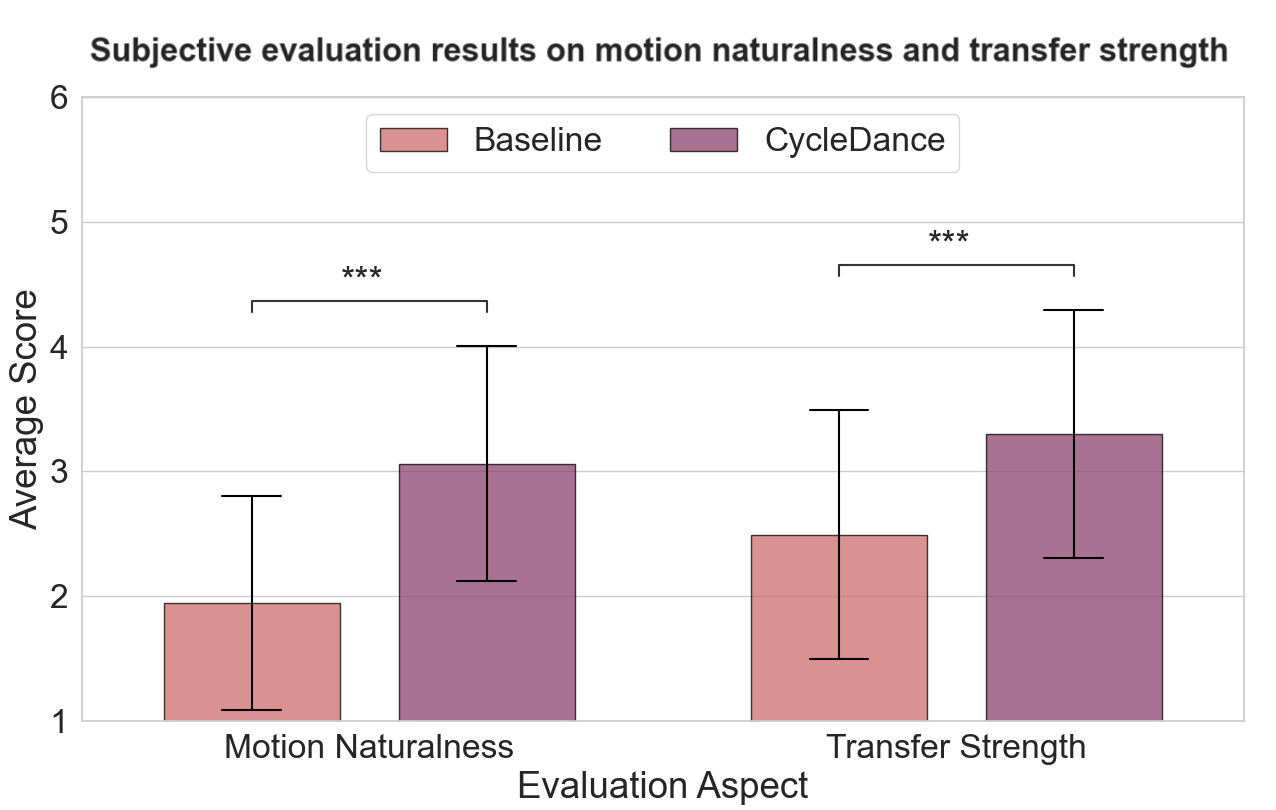}
  \caption{Subjective evaluation results on motion naturalness and transfer strength. Error bars represent standard errors of the averages. Statistical significance is the result of the Wilcoxon signed-rank test that compares the medians ($***$ means $p<0.0001$).}
  \label{fig:sub01}
\end{figure}

\begin{figure}[tp!]
  \centering
  \includegraphics[width=\linewidth]{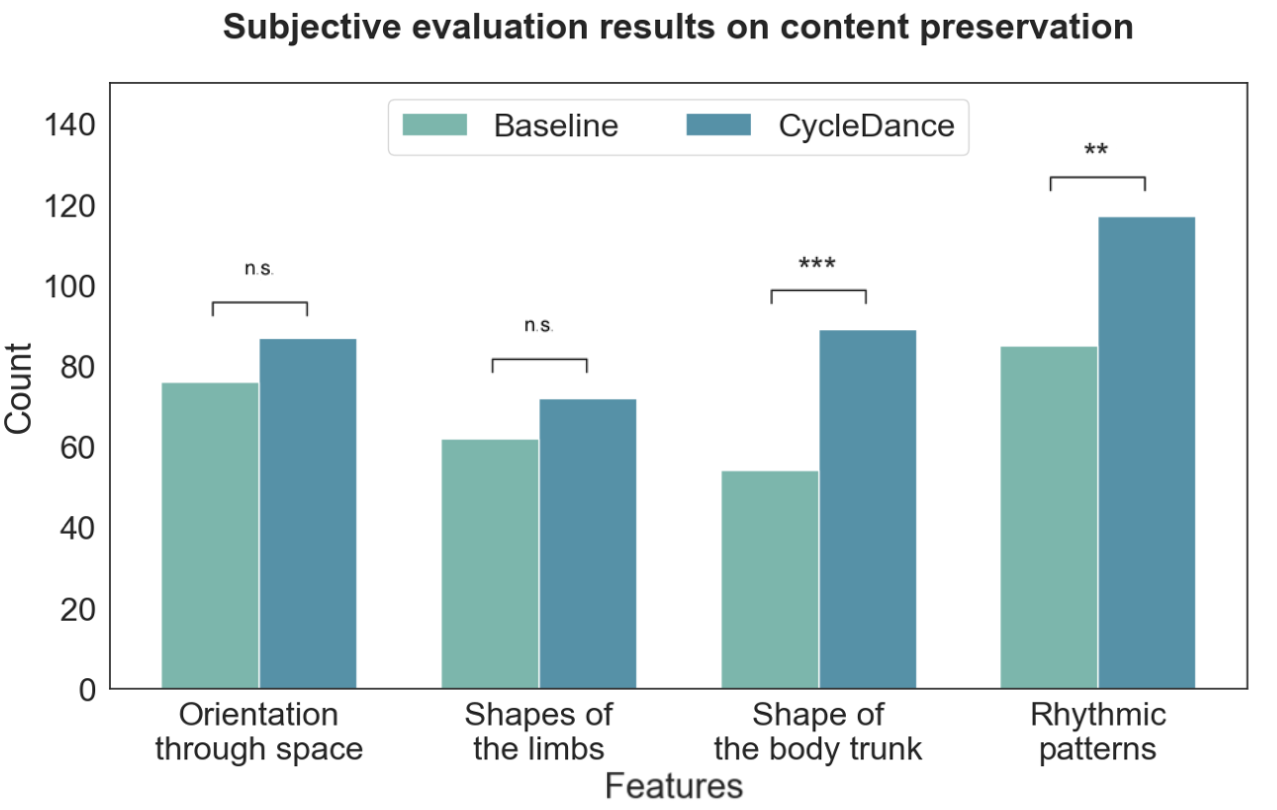}
  \caption{Subjective evaluation results on content preservation. The CycleDance outperforms the baseline model on orientation through space, shape of the limbs, shape of the body trunk, and rhythmic patterns.  Statistical significance represents the results of the Wilcoxon signed-rank test that compares the medians ($***$ means $p<0.00001$, $**$ means $p<0.0001$, and $n.s.$ means $p>0.05$).}
  \label{fig:sub02}
\end{figure}

\begin{figure*}
  \includegraphics[width=\linewidth]{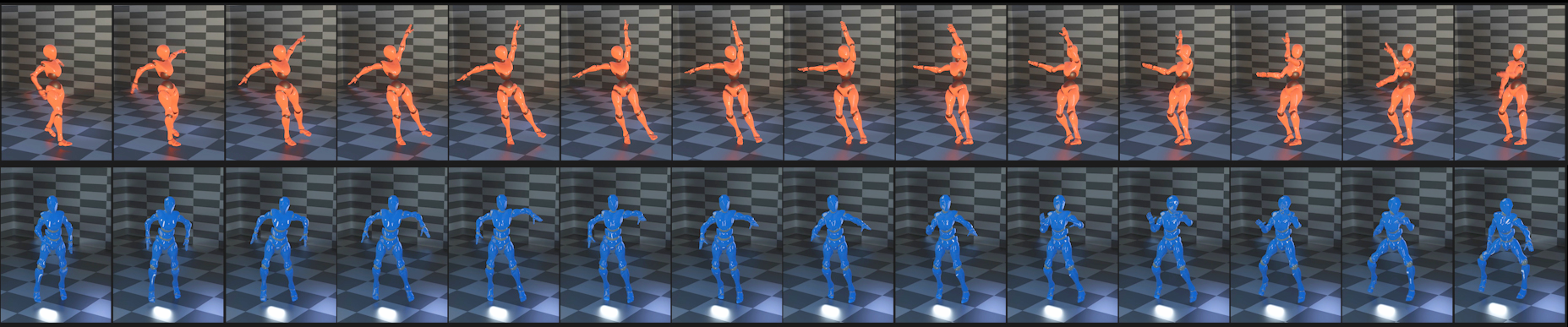}
  \caption{Example ballet-jazz dance (top, red x-bot) transferred to locking dance (bottom, blue y-bot) by CycleDance. }
  \label{fig:sub03}
\end{figure*}

We analyzed the subjective responses to provide statistical support for the results of the user study, and assessed whether the proposed method could be further improved. 
Figure \ref{fig:sub01} demonstrates the responses of the motion naturalness and transfer strength. 
On both aspects, the experts rated CycleDance higher on average compared to the baseline model. 
The subjective responses were compared through a Wilcoxon signed-rank test statistical significance. 
Both the median value of motion naturalness $(Z=-9.2262, p<0.0001)$ and transfer strength $(Z=-8.7677, p<0.0001)$  were significantly higher for CycleDance compared to the baseline model.  
Thus from the view of dance experts, CycleDance is favoured for improved naturalness as well as similarity to the target dance style, consistent with what we observe from the objective quantitative results (Section \ref{sec:objeva}).
As for the responses of the content preservation, Figure \ref{fig:sub02} summarizes the total statistics on the four aspects queried. 
On all four aspects, the experts chose CycleDance more often than the baseline model, when asked which specific features they believe
are preserved. 
We ran a McNemar test for assessing the statistical significance of these gaps. The
test revealed no significant statistical differences between the Baseline model and CycleDance on 'Orientation through space' $(p=0.1724)$ and 'Shapes of the limbs' $(p=0.1573)$. In terms of median value of the 'Shapes of the body trunk' $(p=0.000002)$ and 'Rhythmic patterns' $(p=0.00004)$, 
the McNemar test showed a strong significance in support of the proposed CycleDance model. 
Among the four aspects, both CycleDance and the baseline received higher scores on rhythmic patterns and orientation through space. This implies that it is comparatively easier to keep dance orientation and rhythm while performing dance style transfer. CycleDance outperforms the baseline on preserving the shape of the body trunk. Preserving the shape of the limbs, on the other hand, appears to be more challenging. 
%

Responding to open-ended questions, the dance experts commented that for 'ballet-jazz to locking dance', both methods have a jerky style that emulates pop and lock dance. 
The example shown in Figure \ref{fig:sub03} is frequently mentioned as a major indication of 'transfer' with a visible locking dance style from the view of experienced dancers. 
For CycleDance samples of transferring 'locking dance to ballet-jazz', the dance experts responded that the character arms are clearly jazz or ballet and are really good at holding 'traditional' shapes.
The dance experts also commented on some limitations. One commonly mentioned point is that some motions look wobbly, which may indicate the need for applying some filters to smooth the generated results. The experts also pointed out that ballet-jazz usually requires dancers to point their feet while the generated motions always show flexed ankle joints. This shows the limitation of the considered data which currently do not capture fine foot movements. This caveat also causes some physically unrealistic effects such as the character appears floating when sometimes its body does not have a contact point on the floor. 

\section{Discussion of Societal Impact}
This work contributes a framework for style transfer that aims to offer artistic and scientific innovations to the field of dance. In the short term, we could foresee several impacts on industries and society. 
The positive effect would be the progress in choreographic practice and dance research, which unlock new possibilities in terms of hybrid human-artificial co-creation of dance material. Certain industry sectors could benefit as well, such as video game and animation industries (e.g., group dances where each character has a different motion style). Such effects could
lead to displacement of jobs and a shift towards jobs that relies more on a combination of creativity and automation, as well as development of new user-friendly interfaces and tools.
We also foresee potentially negative impacts or misuses. This technology could blur the lines of ownership in creative processes, i.e., who is/are the creator(s). 
For movement styles beyond dance, such transfer models, if trained on non-representative datasets, could reinforce movement stereotypes of certain societal groups by learning a biased association between group membership and movement styles, e.g.~elderly people or people with disabilities. 

\section{Conclusion and Future work}
\label{sec:conclusion}
\thispagestyle{empty}

This work explores challenging style transfer for sequential data with rich variations and complex frame dependencies such as dance movements. The proposed CycleDance manages to alleviate these challenges by exploiting expressive data encoders, cross-modal contexts and a curriculum based training scheme. Quantitative results from similarity metrics and human expert evaluations confirm the effectiveness of CycleDance. To the authors' knowledge, this is the first work where music context is used for dance or general motion style transfer. 
In the future, we plan to extend the backbone from the CycleGAN-based model to StarGAN or an AdaIN-based model to handle more than two dance genres. Research is also needed to address identified limitations on preserving limb shapes. Based on these techniques, we envision new tools in dance motion design for choreography, film industry, and video games. 

\section*{Acknowledgements}

This study has received funding from the European Commission Horizon 2020 research and innovation program under grant agreement number 824160 (EnTimeMent). This work benefited from access to the HPC resources provided by the Swedish National Infrastructure for Computing (SNIC), partially funded by the Swedish Research Council through grant agreement no. 2018-05973.

{\small

\bibliographystyle{ieee_fullname}
\thispagestyle{empty}
\pagestyle{empty}
\bibliography{egbib}
}
\thispagestyle{empty}
\pagestyle{empty}
\end{document}